\documentclass[letterpaper, 10 pt, conference]{ieeeconf}  

\IEEEoverridecommandlockouts                              

\usepackage{mathptmx} 
\usepackage{times} 
\usepackage{amsmath} 
\usepackage{amssymb}  
\usepackage{mathtools}

\usepackage{graphicx}
\usepackage{siunitx}
\usepackage{multirow}
\usepackage{acro}
\usepackage{url}
\usepackage{hyperref}
\usepackage{cleveref}

\crefname{table}{Table}{Tables}
\Crefname{table}{Table}{Tables}
\crefname{figure}{Fig.}{Figs.}
\Crefname{figure}{Figure}{Figures}

\DeclareAcronym{ma}{
    short = MA,
    long = mechanical advantage,
}
\DeclareAcronym{com}{
    short = COM,
    long = center of mass,
}
\DeclareAcronym{dof}{
    short = DOF,
    long = degree of freedom,
}
\DeclareAcronym{dd}{
    short = DD,
    long = direct-drive,
}
\DeclareAcronym{qdd}{
    short = QDD,
    long = quasi-direct-drive
}
\DeclareAcronym{se}{
    short = SE,
    long = series-elastic,
}
\DeclareAcronym{pe}{
    short = PE,
    long = parallel-elastic,
}
\DeclareAcronym{grf}{
    short = GRF,
    long = ground reaction force,
}
\DeclareAcronym{leap}{
    short = LEAP,
    long = Linear Elastic Actuator in Parallel,
}
\DeclareAcronym{vja}{
    short = VJA,
    long = vertical jumping agility,
}
\DeclareAcronym{imu}{
    short = IMU,
    long = inertial measurement unit,
}
\DeclareAcronym{mcu}{
    short = MCU,
    long = microcontroller unit,
}
\DeclareAcronym{bldc}{
    short = BLDC,
    long = brushless DC,
}
\DeclareAcronym{lipo}{
    short = LiPo,
    long = lithium polymer,
}
\DeclareAcronym{fps}{
    short = FPS,
    long = frames per second,
}

\title{\LARGE \bf
A Long-Legged, Direct-Drive Monopedal Robot Achieves Exceptional Jump Height
}

\author{Gihyeok Na and Justin K. Yim%
\thanks{This work was supported by the National Science Foundation under Grant 2220924.}%
\thanks{The authors are with the Department of Mechanical Science and Engineering, University of Illinois Urbana-Champaign, IL 61801 USA (e-mail: {\tt\small gihyeok2@illinois.edu; jkyim@illinois.edu}).}}%

\begin{document}

\maketitle
\thispagestyle{empty}
\pagestyle{empty}


\begin{abstract}
We demonstrate a jumping robot that reaches high (\SI{7.6}{\meter}) and fast (\SI{190}{\milli\second} stance time) jumps from a single long leg driven by a \acl{dd} transmission, without elastic energy storage. At \SI{281}{\gram}, it achieves the highest jump yet reported for an electrically actuated, spring-free system. The leg uses a new fabric-wrap transmission that provides a variable mechanical advantage, keeping a small electric motor near its peak power output through the stroke while bracing the long, lightweight leg against buckling. A balancing module at the top of the leg uses small propellers to control the leg's orientation on the ground and in the air, where the long leg provides a large moment arm for the control torques. The robot is validated outdoors with vertical jumps, attitude control on the ground and in flight, and tilted jumps.
\end{abstract}

\section{INTRODUCTION}
\subsection{Motivation}\label{subsec:motivation}
Jumping is an exciting motion that enables terrestrial locomotors to navigate rugged natural environments and complicated built environments. In particular, jumping can enable fast motion, rapid changes in direction, long ballistic flight phases, and generation of large impulsive forces with peaks greater than a system's weight. Locomotors in both biology and robotics deploy jumping to tackle a variety of challenges including scaling obstacles or ledges~\cite{stoeter2002autonomous, klemm_ascento_2019}, transitioning between locomotion on surfaces of different slope~\cite{haldane_robotic_2016, xu2025pinto}, crossing gaps between footholds~\cite{yang2025agile}, escaping predation or capturing prey, and more. Jumping has even been considered for extraterrestrial locomotion, where lower gravity could uniquely enable small jumping rovers to rapidly traverse challenging terrain on small bodies~\cite{hockman2017design, hockman2022gravity, wagner_underactuated_2026}. Robotic jumping research has addressed a variety of performance metrics: jumpers surmount obstacles better than rolling platforms~\cite{hougen2000miniature, ackerman_boston_2012, klemm_ascento_2019}; offer faster direction changes, higher payload capacity, and longer endurance than flying platforms~\cite{zhu2022pogodrone, wang2024terrestrial, bai_agile_2024}; and reach exceptional efficiency when leveraging elastic energy storage~\cite{guenther2016energy}.

Jump height is one of the most fundamental jump performance metrics and it has been compared in several ways. Normalized by body length, jump height scales weakly with size, so smaller jumpers achieve much higher body-length-normalized jumps~\cite{kovac_miniature_2008}. Jump height has also been scaled by power and time, particularly in the context of hopping, where consecutive jumps follow one another in a bouncing motion.

    \begin{figure}[tbp]
        \centering
        \includegraphics[width=.7\columnwidth]{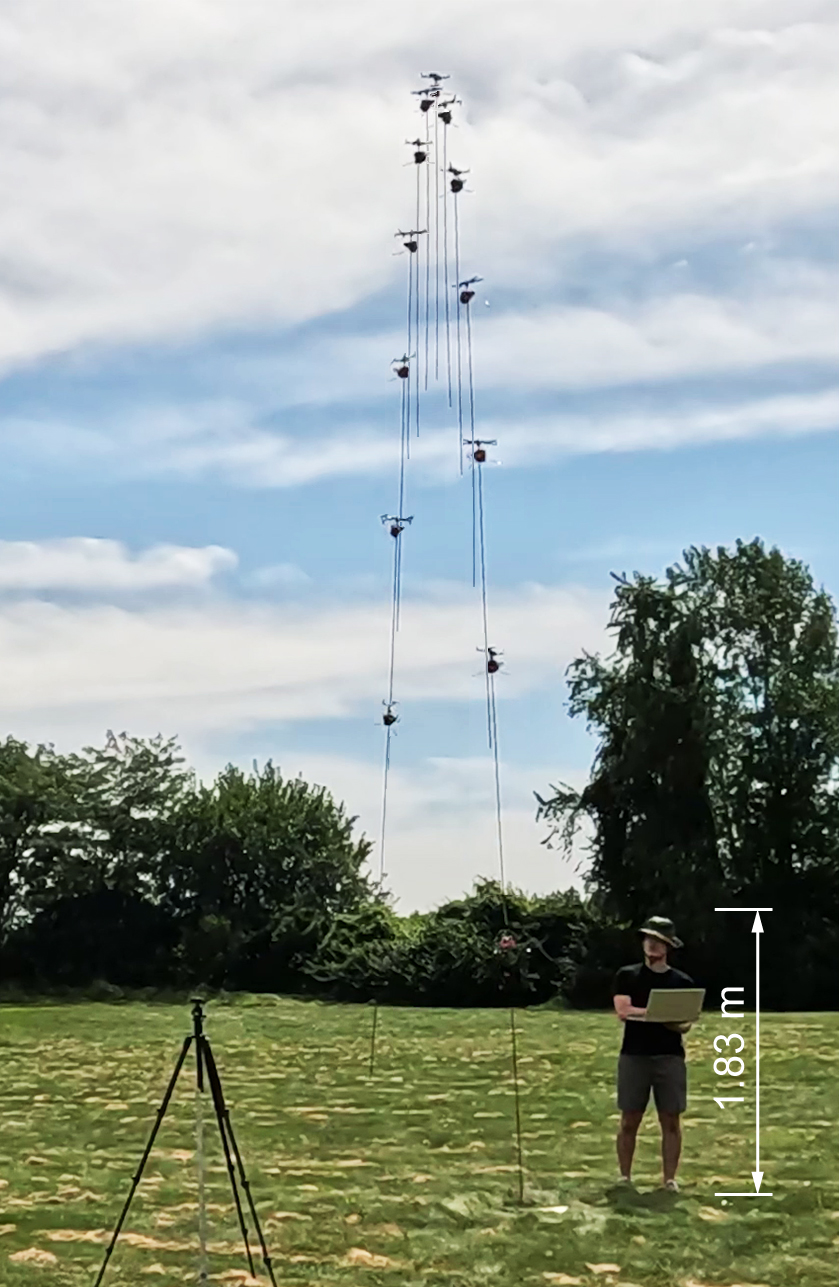}
        \caption{Overlay of a single vertical jump, composited from frames \SI{166.7}{\milli\second} apart, with a person of height \SI{1.83}{\meter} for scale.}
        \label{fig:vertical_jump_overlay}
        \vspace{-4mm}
    \end{figure}

Following~\cite{haldane_robotic_2016}, we characterize jumping performance by two quantities, jump height and jump frequency. We take the jump height $h$ as the vertical displacement of the \ac{com} from motion start to the apex. The jump frequency is defined as
\begin{equation}
    f = \frac{1}{t_s + t_f}\quad(\unit{\hertz}),
    \label{eq:jump_freq}
\end{equation}
where $t_s$ is the stance duration in ground contact and $t_f$ is the flight time from liftoff to the apex. \Ac{vja} is the jump height multiplied by the jump frequency,
\begin{equation}
    \mathrm{VJA} = hf = \frac{h}{t_s + t_f}\quad(\unit{\meter\per\second}),
    \label{eq:vja}
\end{equation}
which can be read as the maximum rate at which a system could ascend by hopping up steps spaced by its jump height. A related dimensionless metric is the normalized hopping frequency~\cite{bai_parallel-elastic_2026}.

Increasing a jumper's jump height expands its unique ability to cross large obstacles with ballistic motions. Robots that also score highly on \ac{vja}, by producing rapid, powerful jumps, can additionally traverse their environment quickly.

\subsection{Contribution}\label{subsec:contributions}
\begin{figure}[tpb]
        \centering
        \includegraphics[width=.7\columnwidth]{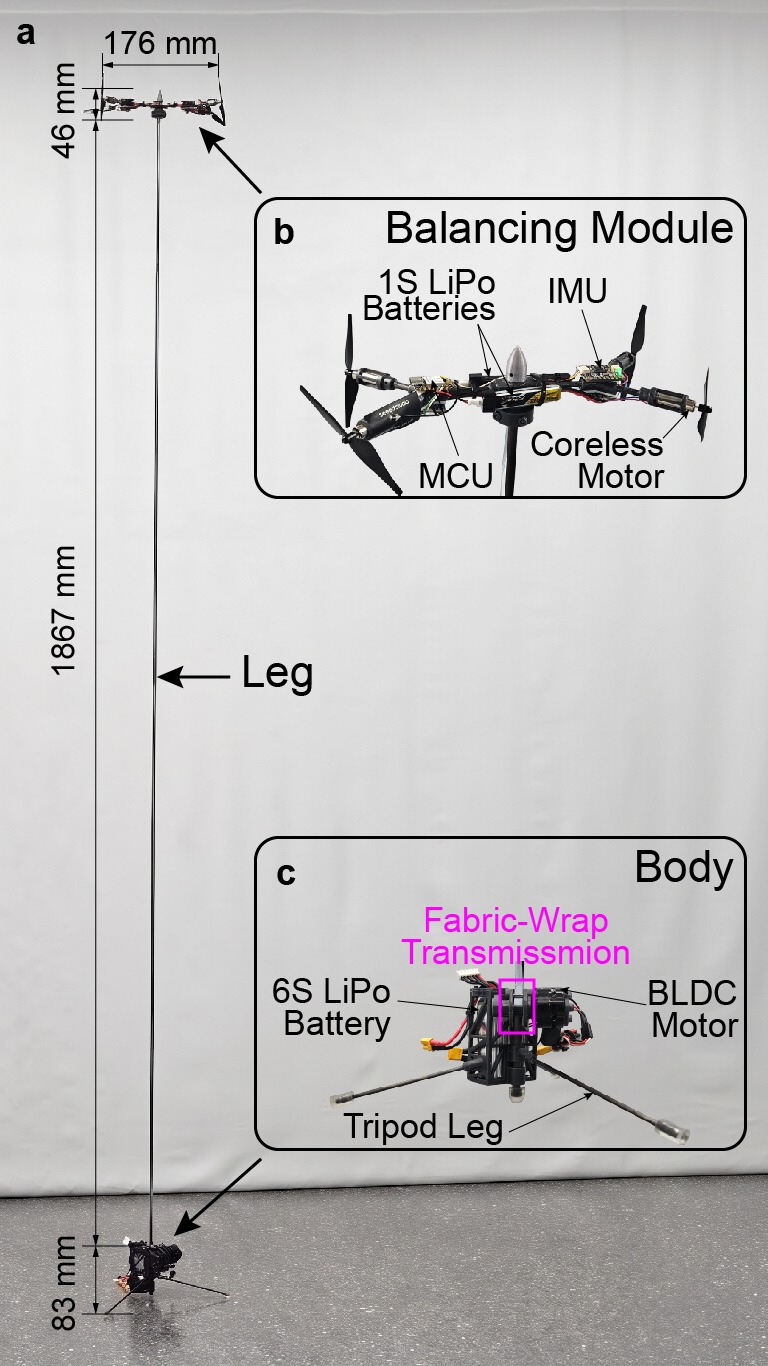}
        \caption{Overview of the robot and its three subsystems: the body, the balancing module, and the leg. (a)~The full robot, with a leg stroke of \SI{1867}{\milli\meter} and a body height of \SI{83}{\milli\meter}. (b)~The balancing module, mounted at the top of the leg. (c)~The body, containing the main actuator.}
        \vspace{-4mm}
    \label{fig:system_overview}
    \end{figure}

We introduce the fabric-wrap transmission, a direct-drive mechanism that pairs a long leg stroke with a variable \ac{ma}. Built into a monopedal robot, it achieves an exceptional jump height for an electrically actuated system, which we validate in a single jump hardware experiment (\cref{fig:vertical_jump_overlay} and \cref{fig:system_overview}).
    
    The main contributions of this work are:
    \begin{itemize}
        \item a hardware demonstration reaching a jump height of \SI{7.6}{\meter}
        \item introduction of the fabric-wrap transmission, which provides a variable \ac{ma} through the stroke without added gearing
        \item integration of a balancing module that enables attitude control and tilted jumps
    \end{itemize}

\subsection{Jumping Actuation Methods}\label{subsec:jumping_actuation_methods}
\begin{figure}[tbp]
        \centering
        \includegraphics[width=.7\columnwidth]{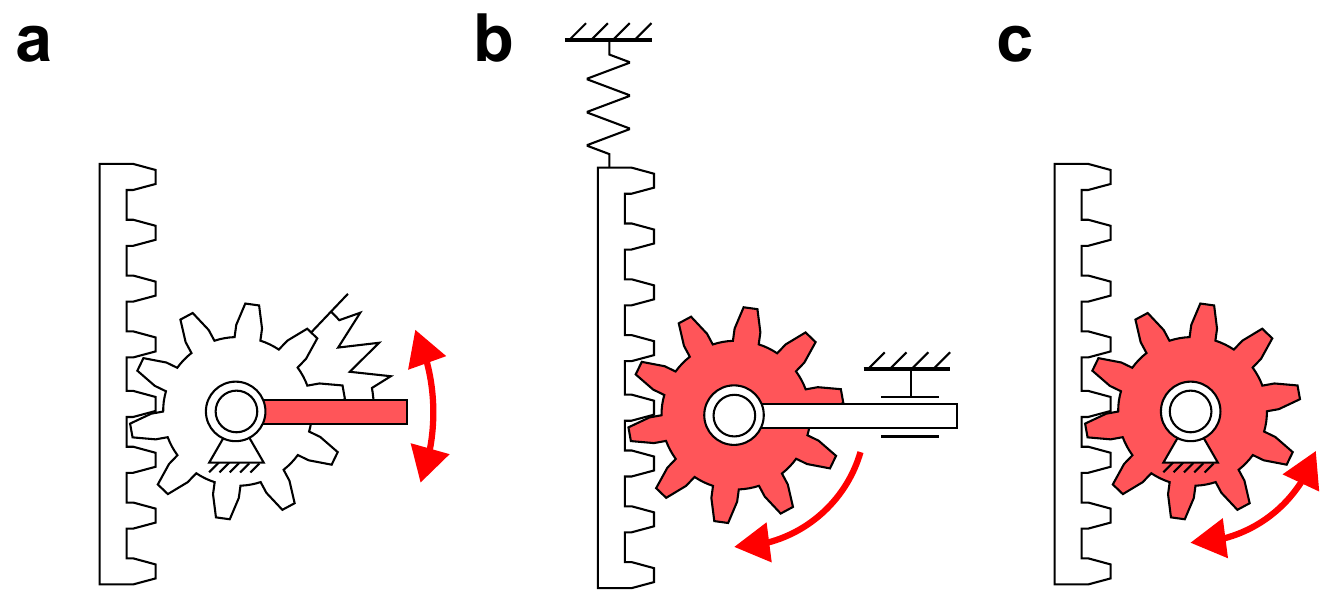}
        \caption{Actuation strategies, each drawn as a rack and pinion. The colored part is the actuated element, the rack is the leg, and the red arrow marks the actuator output. (a)~\Acf{se} actuation: elastic element in series. (b)~\Acf{pe} actuation: elastic element in parallel, with a latch. (c)~\Acf{dd} actuation: rigid coupling, no elastic element.}
        \vspace{-4mm}
    \label{fig:actuation_methods}
    \end{figure}

The actuation methods of jumping robots are broadly classified into three types: \ac{se}, \ac{pe}, and \acf{dd} actuations (\cref{fig:actuation_methods}).

\Ac{se} systems place a spring between the actuator and the leg (\cref{fig:actuation_methods}~(a)). The spring stores energy through stance and releases it rapidly, giving an instantaneous output power beyond what the actuator can deliver directly~\cite{haldane_robotic_2016}. It can vary jump energy from stride to stride in continuous hopping at adjustable height, and the spring shields the actuator from impact loads. The costs are added mechanical complexity and usually a longer stance, since the leg must stay grounded to charge the spring before each jump.

\Ac{pe} systems place a spring in parallel with the leg, loading it through a large reduction so that even a small, low-power motor can store substantial energy (\cref{fig:actuation_methods}~(b)). A latch holds the spring against back-driving, and a trigger releases it. This achieves the highest reported jump heights~\cite{hawkes_engineered_2022}, at the cost of additional winding and triggering mechanisms and increased complexity in modulating energy mid-jump~\cite{divi2020latch}.

\Ac{dd} systems couple the actuator rigidly to the leg, with no elastic element in the powertrain, so the ground force tracks the actuator output directly and the jump power cannot exceed the actuator's instantaneous limit (\cref{fig:actuation_methods}~(c)). In practice, \ac{dd} actuation is often realized as \ac{qdd} actuation, with a small back-drivable reduction (typically 4:1 to 10:1) that raises force while preserving the elastic-free coupling; we treat the two together. Without air drag, the jump height of a power-limited actuator scales as
\begin{equation}
    h_{\max} \propto \left(l_\text{leg}\frac{P_{\max}}{M}\right)^{2/3},
    \label{eq:DD_jump_height}
\end{equation}
where $M$ is the system's total mass, $l_\text{leg}$ the leg stroke, and $P_{\max}$ the actuator's peak power~\cite{bai_parallel-elastic_2026}. This scaling sets our core inspiration: rather than add elastic elements, we keep the minimal \ac{dd} powertrain and maximize its two remaining levers, leg stroke and power-to-mass ratio.

\Ac{dd} actuation brings three further advantages. Removing the spring, winding, and trigger yields a simpler, lighter powertrain and a shorter stance, since nothing must be charged before a jump. Rigid coupling lets the \ac{grf} be commanded continuously through stance and gives the force loop high bandwidth, since no series compliance filters the actuator output. Finally, leg force can be estimated from motor current alone, which could enable proprioceptive sensing without a dedicated sensor.

\subsection{Related Works}\label{subsec:related_works}
\Ac{se} and \ac{pe} designs have been used in several jumpers, especially small-sized ones. Salto and Salto-1P reach a standing vertical leap of \SI{1.25}{\meter} from a leg length of only \SI{14.4}{\centi\meter} and a mass of \SI{98}{\gram} using series-elastic power modulation~\cite{haldane_robotic_2016, haldane2017repetitive}. Ascento, a bipedal wheeled-legged robot, reached a maximum jump height of \SI{0.4}{\meter} with a total mass of \SI{10.4}{\kilo\gram}~\cite{klemm_ascento_2019}.

\Ac{pe} design has also been widely used for jumping robots, and has produced the highest jump heights reported to date. Hawkes et al. demonstrated a jumper reaching \SI{32}{\meter}~\cite{hawkes_engineered_2022}. This uses a very high gear reduction (1000:1) and a latch to store a substantial amount of energy in its elastic elements, but must remain on the ground for approximately \SI{120}{\second} to recharge between jumps. Bai et al. recently achieved a hopping height of \SI{3.6}{\meter} with high continuity, using a reactive latch to extend actuation over the aerial phase~\cite{bai_parallel-elastic_2026}. They also demonstrated a quadcopter-driven hopper with a passive telescopic leg, reaching \SI{1.63}{\meter} using rotor thrust rather than leg actuation to control jump height and continuity~\cite{bai_agile_2024}. JumpRoACH uses torsional springs and latex rubber bands to achieve a \SI{1.5}{\meter} jump~\cite{jung_jumproach_2019}. Kovac et al. introduced a \SI{5}{\centi\meter}, \SI{7}{\gram} robot that achieves a jump height of \SI{1.4}{\meter}, more than 27 times its own body size, using two parallel torsion springs~\cite{kovac_miniature_2008}. The \ac{leap} mechanism uses \ac{pe} design at a similarly small scale~\cite{batts_design_2016, kulic_untethered_2017}. These results show that elastic energy storage can reach substantial jump heights, but only with the added winding, latch, or trigger mechanisms described in \cref{subsec:jumping_actuation_methods}.

\Ac{dd} actuation drives several quadruped robots. Stanford Doggo, an open-source \ac{qdd} quadruped, reaches a vertical jump of \SI{1.14}{\meter}~\cite{kau_stanford_2019}, while Minitaur's \ac{dd} legs enable a range of dynamic behaviors~\cite{kenneally_design_2016}. The MIT Cheetah 2 demonstrated a highly efficient \ac{qdd} system~\cite{park_variable-speed_2015, park_high-speed_2017, park_jumping_2021}. These platforms prioritize general locomotion, so their relatively short leg stroke and high mass yield modest jump heights.

Closer to our goal are \ac{dd} monopedal jumpers built specifically to maximize height. RAMIEL, a parallel-wire-driven monoped, achieves a \SI{1.6}{\meter} jump using \ac{qdd} winding modules and a lightweight leg, pursuing a long, low-mass leg driven through a wire-based transmission~\cite{suzuki_ramiel_2022}. Wagner and Yim demonstrated a multimodal robot with a single leg driven by a rack-and-pinion transmission~\cite{wagner_underactuated_2026}. The leg itself serves as the rack, driven directly by a pinion gear coupled to the motor, keeping the powertrain simple, as shown in \cref{fig:fabric_wrap}~(d). This robot reaches a jump height of \SI{0.59}{\meter}. Unlike most legged robots, neither of these monopedal robots relies on a linkage mechanism for its leg. Instead, both keep a minimal, rod-like leg, similar to this work.

\begin{table}[tbp]
\caption{Comparison of Electrically Actuated Jumping Robots}
\label{tab:dd_comparison}
\vspace{-7mm}
\begin{center}
\footnotesize
\setlength{\tabcolsep}{3pt}
\begin{tabular}{llcccc}
\hline
\multirow{2}{*}{Type} & \multirow{2}{*}{Reference} & Mass & Leg stroke & Jump height & \acs{vja}\\
 & & (\SI{}{\kilo\gram}) & (\SI{}{\centi\meter}) & (\SI{}{\meter}) & (\SI{}{\meter\per\second})\\
\hline
\multirow{6}{*}{\acs{dd}} & \textbf{This work} & \textbf{0.28} & \textbf{187} & \textbf{7.6} & \textbf{5.30}\\
 & Wagner and Yim~\cite{wagner_underactuated_2026} & 1.25 & 20 & 0.59 & 1.19\\
 & Suzuki et al.~\cite{suzuki_ramiel_2022} & 10.3 & 80 & 1.6 & 2.00\\
 & Kau et al.~\cite{kau_stanford_2019} & 4.8 & 16 & 1.14 & 2.23\\
 & Park et al.~\cite{park_variable-speed_2015, park_high-speed_2017, park_jumping_2021} & 33 & 80 & 0.5 & 1.11\\
 & Kenneally et al.~\cite{kenneally_design_2016} & 5 & 20 & 0.48 & 1.12\\
\hline
\multirow{2}{*}{\acs{se}} & Haldane et al.~\cite{haldane2017repetitive} & 0.098 & 14.4 & 1.25 & 1.83\\
 & Klemm et al.~\cite{klemm_ascento_2019} & 10.4 & 35 & 0.4 & 0.92\\
\hline
\multirow{4}{*}{\acs{pe}} & Hawkes et al.~\cite{hawkes_engineered_2022} & 0.03 & 20.3 & 32 & 0.25\\  
 & Jung et al.~\cite{jung_jumproach_2019} & 0.099 & 9.5 & 1.5 & 0.03\\
 & Kovac et al.~\cite{kovac_miniature_2008} & 0.007 & 5 & 1.4 & 0.35\\
 & Bai et al.~\cite{bai_parallel-elastic_2026} & 0.1 & 15 & 3.6\textsuperscript{*} & 3.93\\
\hline
\end{tabular}
\end{center}
{\footnotesize\textsuperscript{*}The reported jump height includes additional height from quadcopter thrust.}
\vspace{-2mm}
\end{table}

\Cref{tab:dd_comparison} compares several of these electrically actuated robots on mass, leg stroke, jump height, and \ac{vja}.

\section{METHODS}
\subsection{System Overview}\label{subsec:system_overview}
The robot consists of three subsystems: the leg, the body, and the balancing module, shown in \cref{fig:system_overview}. The leg is a long carbon-fiber tube that forms the robot's main structure, the body sits at its lower end and houses the jump actuator and transmission, and the balancing module is mounted at the upper end and provides attitude control. We designed the body and the balancing module so that each has its \ac{com} close to the leg axis, placing the combined \ac{com} on the axis and simplifying attitude stabilization.

The leg tube measures \SI{1981}{\milli\meter} in length, with outer and inner diameters of \SI{6}{\milli\meter} and \SI{5}{\milli\meter} and a mass of \SI{27}{\gram}. The low mass keeps the tube's share of the total system mass small, but its correspondingly low bending stiffness makes the first bending mode easy to excite: with the balancing module active and the leg in ground contact, the tube vibrates in this mode unless the body's tripod foot constrains the base.

The body carries a Vertiq 23-06 2200~Kv \ac{bldc} motor as the main actuator, powered by a 6S \ac{lipo} battery (\SI{550}{\milli\ampere\hour}, 95C). The motor drives the fabric-wrap transmission by spinning the spool: one end of the fabric is anchored near the top of the leg just below the balancing module and the other winds onto the spool, so motor torque becomes fabric tension and, in turn, the vertical \acf{grf} that drives the jump. The fabric is a commercial Dyneema\textregistered{} composite hybrid (CT5K.18/wov.32c, 2.92\,oz/yd\textsuperscript{2}), \SI{6}{\milli\meter} wide and \SI{0.13}{\milli\meter} thick, chosen for its high tensile strength so that it does not fail under the jump loads. We detail the mechanism and the fabric selection in \cref{subsec:fabric_wrap_transmission}. An ESP32 \ac{mcu} drives the motor, and a compliant carbon-fiber tripod at the base supports the leg and suppresses the vibration when the balancing module is active.

The balancing module, mounted at the top of the leg, carries four propeller thrusters, its own \ac{mcu} and \ac{imu}, and two 1S \ac{lipo} batteries (\SI{220}{\milli\ampere\hour} and \SI{300}{\milli\ampere\hour}). It controls the robot's orientation during both stance and flight, and is described in \cref{subsec:balancing_module}.

\begin{table}[tbp]
\caption{System Mass Breakdown}
\label{tab:mass}
\vspace{-4mm}
\begin{center}
\footnotesize
\setlength{\tabcolsep}{3pt}
\begin{tabular}{llc}
\hline
Subsystem & Component & Mass (\SI{}{\gram})\\
\hline
\multirow{4}{*}{Body}  & Motor & 53\\
& Battery & 88\\
 & Chassis, electronics, wiring & 51\\
\cline{2-3}
 & Subtotal & 192\\
\hline
\multirow{4}{*}{Balancing module} & Motors ($4\times$) & 12\\
 & Batteries ($2\times$) & 14\\ 
 & Chassis, electronics, wiring, props & 29\\
\cline{2-3}
 & Subtotal & 55\\
\hline
\multirow{2}{*}{Leg} & Carbon-fiber tube & 27\\
 & Others & 7\\
\cline{2-3}
 & Subtotal & 34\\
\hline
\textbf{Total} & & \textbf{281}\\
\hline
\end{tabular}
\end{center}
\vspace{-4mm}
\end{table}

\Cref{tab:mass} lists the mass of each subsystem. The total system mass is \SI{281}{\gram}.

\subsection{The Fabric-Wrap Transmission}\label{subsec:fabric_wrap_transmission}
\begin{figure}[tbp]
        \centering
        \includegraphics[width=.75\columnwidth]{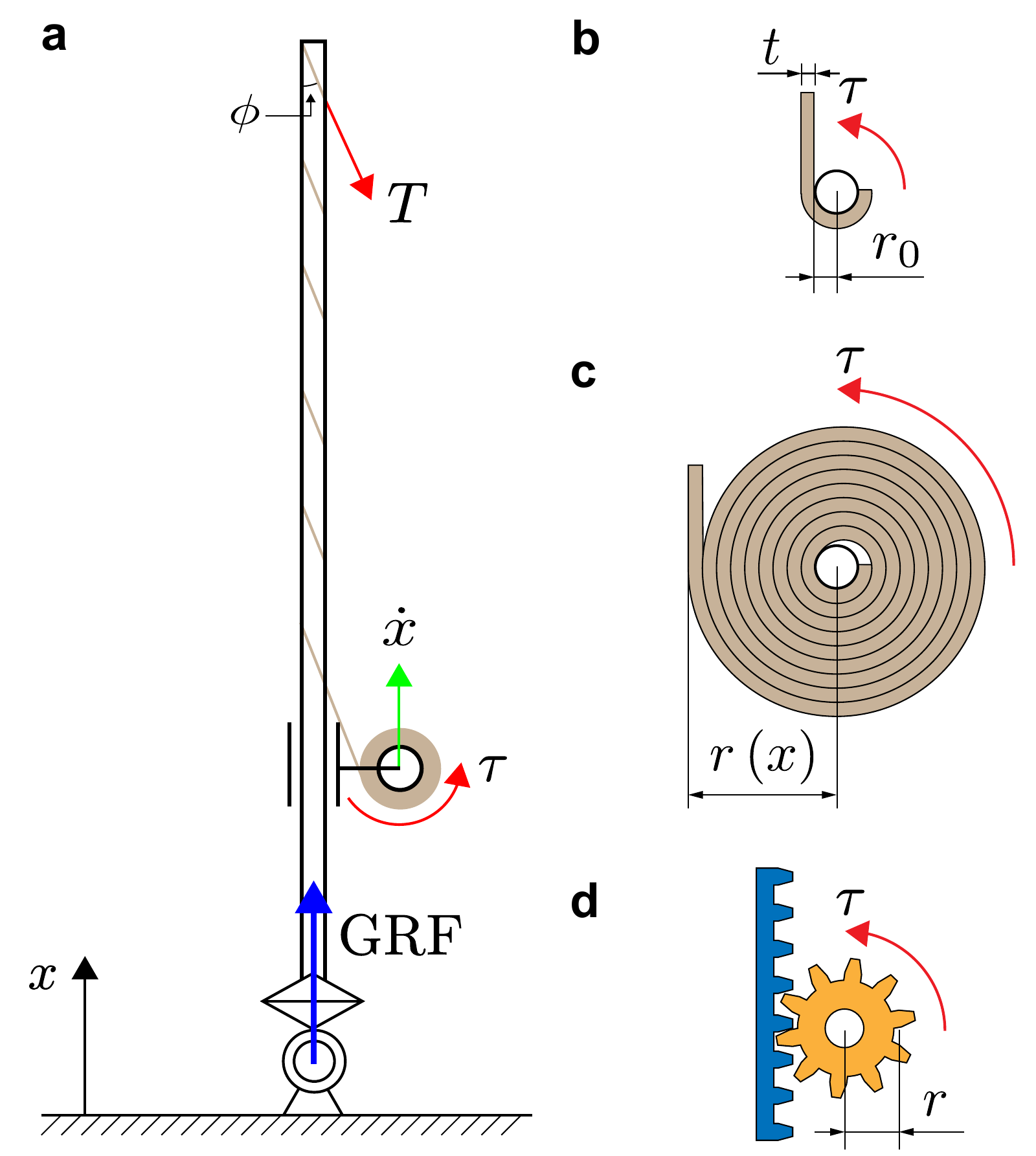}
        \caption{Schematic of the fabric-wrap transmission principle. (a)~The fabric spirals around the leg at a spiral wrap angle $\phi$, converting motor torque $\tau$ into tension $T$ and the vertical \acf{grf} that drives the motion $x$ along the leg. (b)~At the start of stance, the fabric wraps at radius $r_0$. (c)~As the leg extends, the radius grows to $r(x)$, decreasing the \acf{ma}. (d)~A rack and pinion~\cite{wagner_underactuated_2026}, by contrast, keeps $r$ fixed, causing a constant \ac{ma}.}
        \vspace{-4mm}
        \label{fig:fabric_wrap}
    \end{figure}

The fabric-wrap transmission, the core mechanism of this work, consists of a fabric tape and a spool. Its principle is illustrated in \cref{fig:fabric_wrap}.

A fabric tape wraps helically around the leg rod at a fixed angle $\phi$ from the leg axis and winds onto a motor-driven spool, converting motor torque into leg force (\cref{fig:fabric_wrap}~(a)). The wrap serves two further purposes: it prevents the long, thin leg tube from buckling under the ground reaction force during a jump, and it damps the tube's bending vibration when the balancing module is active.

As the motor winds in fabric of thickness $t$, the spool radius grows from its initial value $r_0$ with the leg extension $x$ (\cref{fig:fabric_wrap}~(b,~c)). A wound length $l$ adds a fabric cross-sectional area $lt = \pi[r\left(x\right)^2 - r_0^2]$ to the spool and advances the body along the leg by $x = l\cos\phi$. Combining these gives the spool radius
\begin{equation}
    r\left(x\right) = \sqrt{r_0^2 + \frac{t}{\pi\cos\phi}\,x}.
    \label{eq:wrap_radius}
\end{equation}
With no slip at the spool, its angular velocity $\dot\theta$ and the body's linear velocity $\dot x$ are related through a variable \ac{ma},
\begin{equation}
    \dot\theta = \mathrm{MA}(x)\,\dot x, \quad \mathrm{MA}(x) = \frac{1}{r\left(x\right)\cos\phi}.
    \label{eq:ma}
\end{equation}

The main actuator is modeled as a DC motor with a firmware current cap. Because the rotor and spool share a shaft, the motor turns at the spool speed $\dot\theta$, and its winding current and delivered torque are
\begin{equation}
    I = \min\!\left(\frac{V_\text{s} - \dot\theta/K_v}{R},\; I_{\max}\right), \quad \tau = K_t I - k_d\,\dot\theta,
    \label{eq:motor}
\end{equation}
where $V_\text{s}$ is the supply voltage, $K_v$ and $K_t$ the velocity and torque constants, $R = R_\text{motor} + S\,R_\text{batt}$ the total series resistance for $S$ cells, $k_d$ a viscous coefficient, and $I_{\max}$ the current limit. The cap sets two regimes. Below the knee speed $\dot\theta_\text{knee} = K_v\left(V_\text{s} - I_{\max} R\right)$, the current saturates at $I_{\max}$, so torque is nearly constant and output power rises linearly with speed. Above it, the current falls and torque decreases with speed in the usual manner. The growing \ac{ma} holds the motor in this current-limited, high-power regime over most of the stroke.

A fixed-ratio drive such as a rack and pinion (\cref{fig:fabric_wrap}~(d)) has constant $r$, so $\dot\theta = \dot x / r$ and the motor speed tracks the body speed directly, sweeping the motor across its full speed range in a single stroke. In the fabric-wrap transmission, by contrast, $r(x)$ grows over the stroke, so the \ac{ma} is large at the start of stance and decreases as the leg extends. This keeps the motor near its peak-power operating range over a larger fraction of the stroke, without an additional gear train.

Because the constraint in \eqref{eq:ma} is holonomic, the rotor, leg, and translating body form a single \ac{dof} system, parameterized by the leg extension $x$. The leg is driven by the same no-slip wrap but at its own fixed outer radius $r_\text{leg}$, so its angular velocity follows a constant ratio
\begin{equation}
    \dot\theta_\text{leg} = \mathrm{MA}_\text{leg}\,\dot x, \quad \mathrm{MA}_\text{leg} = \frac{1}{r_\text{leg}\cos\phi}.
    \label{eq:ma_leg}
\end{equation}
$\mathrm{MA}_\text{leg}$ is constant, as $r_\text{leg}$ is fixed, whereas $\mathrm{MA}$ varies with $x$. The system's kinetic energy $\tfrac{1}{2}m\dot x^2 + \tfrac{1}{2}J\dot\theta^2 + \tfrac{1}{2}J_\text{leg}\dot\theta_\text{leg}^2 = \tfrac{1}{2}(m + J\,\mathrm{MA}^2 + J_\text{leg}\,\mathrm{MA}_\text{leg}^2)\dot x^2$ gives an effective translating mass
\begin{equation}
    M_\text{eff}(x) = m + J\,\mathrm{MA}^2 + J_\text{leg}\,\mathrm{MA}_\text{leg}^2,
    \label{eq:m_eff}
\end{equation}
where $m$ is the mass translating along the leg, $J = J_\text{rotor} + J_\text{spool}$ is the combined rotor and spool inertia, and $J_\text{leg}$ is the leg's moment of inertia about its axis. In this design the leg term is small and constant, so $M_\text{eff}(x)\approx m + J\,\mathrm{MA}^2$. The Euler--Lagrange equation, with the generalized motor force $\tau\,\mathrm{MA}$ and gravity, gives the stance dynamics
\begin{equation}
    \ddot x = \frac{\tau\,\mathrm{MA} - J\,\mathrm{MA}\,\mathrm{MA}'\,\dot x^2 - mg - F_d}{M_\text{eff}(x)},
    \label{eq:eom}
\end{equation}
where $\mathrm{MA}' = d\mathrm{MA}/dx$ and $F_d$ is the aerodynamic drag.

During stance, only the body translates. When it reaches the end of the stroke with velocity $\dot x_f$, it engages the leg and balancing module, and the system leaves the ground as one. Conservation of momentum sets the \ac{com} liftoff velocity
\begin{equation}
    \dot Z_\text{LO} = \frac{m}{M}\,\dot x_f,
    \label{eq:liftoff}
\end{equation}
where $M$ is the total mass of the robot.

After liftoff, the robot follows a ballistic trajectory with aerodynamic drag that we model in simulation as:
\begin{equation}
    \ddot Z = -g - \tfrac{C_d \rho A|\dot Z|}{2M} \dot Z,
    \label{eq:aerial}
\end{equation}
with $\rho$ the air density, $A$ the frontal area, and $C_d$ the drag coefficient. Integrating from the liftoff velocity $\dot Z_\text{LO}$ gives the apex height~\cite{bai_parallel-elastic_2026}
\begin{equation}
    h = \frac{M}{C_d \rho A}\ln\left(\frac{C_d \rho A}{2M}\frac{\dot Z_\text{LO}^2}{g} + 1\right).
    \label{eq:jump_height}
\end{equation}

Because the \ac{ma} governs the body acceleration and hence the liftoff velocity, we chose the transmission geometry by maximizing the predicted jump height over the initial spool radius $r_0$ and the fabric thickness $t$. In simulations, we swept the spiral wrap angle $\phi$ from \SI{1}{\degree} to \SI{6}{\degree} in \SI{1}{\degree} steps, bounding $r_0$ to \SIrange{1}{5}{\milli\meter} and $t$ to \SIrange{0.05}{1}{\milli\meter}. The optimum was nearly independent of $\phi$, at $r_0^* = \SI{1.00}{\milli\meter}$ and $t^* = \SI{0.14}{\milli\meter}$. Since $r_0^*$ sat at the lower bound, a smaller initial radius is always favored. We instead set $r_0 = \SI{2.00}{\milli\meter}$ for manufacturability and durability, with $t = \SI{0.13}{\milli\meter}$, the nearest commercially available thickness, and a spiral wrap angle $\phi = \SI{2.5}{\degree}$. These values depart slightly from optimal with negligible loss in predicted jump height.

\subsection{The Balancing Module}\label{subsec:balancing_module}
The long leg also enables attitude control with low-force actuators. The balancing module sits at the top of the leg, so its thrusters act at a long moment arm about the foot. The module uses four propeller thrusters, each a \SI{7}{\milli\meter}-diameter coreless brushed DC motor fitted with a \SI{76.3}{\milli\meter}-diameter propeller and driven bidirectionally by a motor driver. It also carries an ESP32 \ac{mcu} that communicates wirelessly with the body, a BNO085 \ac{imu} with integrated orientation estimation, and two 1S \ac{lipo} batteries.

    \begin{figure}[tbp]
        \centering
        \includegraphics[width=.75\columnwidth]{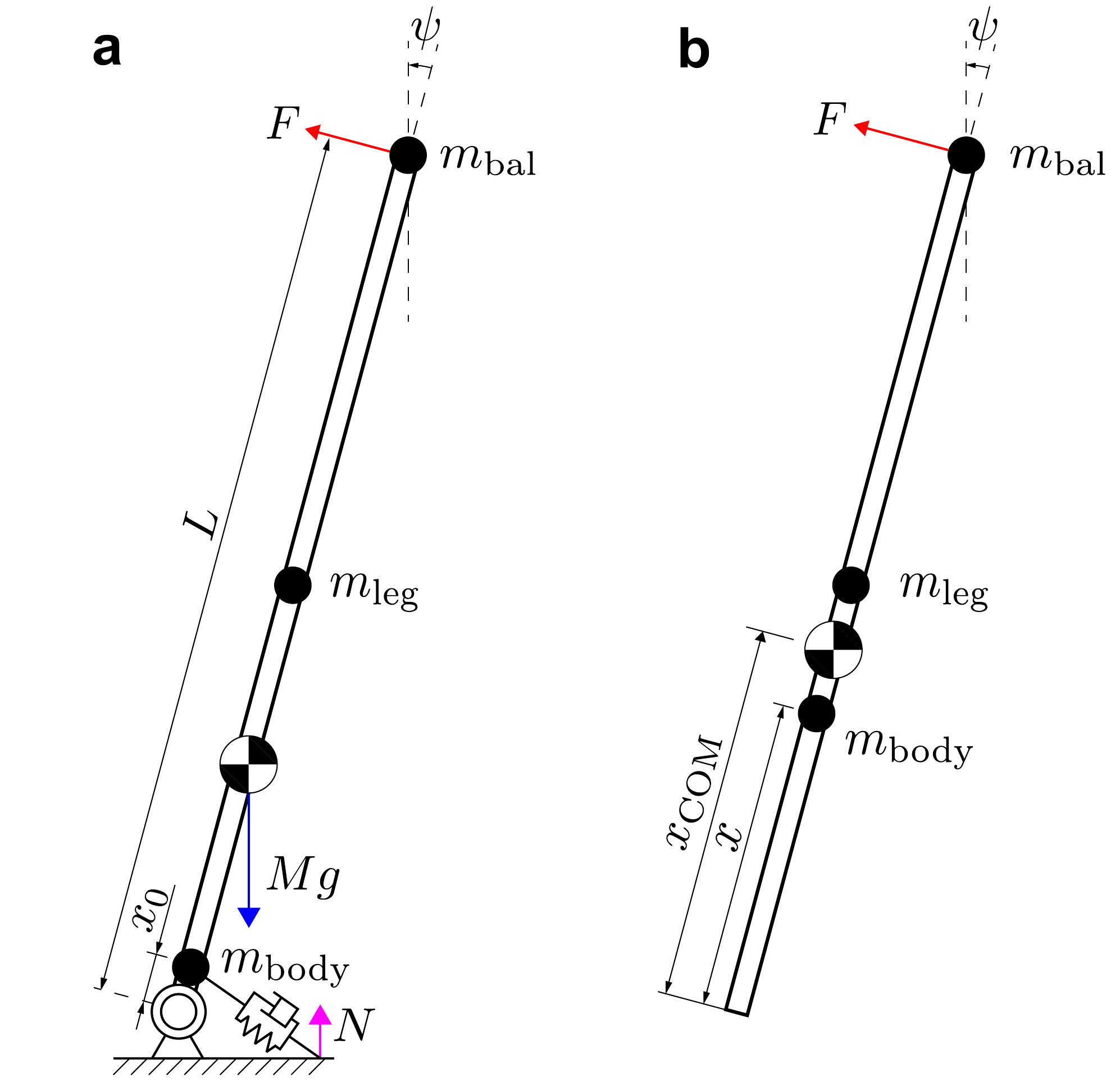}
        \caption{Attitude-control model. $L$ is the length from foot to the balancing module, $F$ is the net thrust at the module, and $\psi$ is the lean from vertical. (a)~Contact phase: the system is an inverted pendulum about the foot, with the body \ac{com} at $x_0$ and a tripod contacting the ground for support. (b)~Aerial phase: the system rotates about its center of mass at $x_\text{COM}$, with the body at $x$. }
        \vspace{-4mm}
        \label{fig:balancing_diagram}
    \end{figure}

The four thrusters are carried on two mounts (\cref{fig:system_overview}~(b)), each holding two thrusters \SI{45}{\degree} apart and offset laterally from the leg axis, so that opposing pairs produce roll and pitch torques. Each thruster produces up to \SI{0.245}{\newton} forward and \SI{0.108}{\newton} in reverse, so the largest net force is $2(0.245+0.108)\cos\SI{45}{\degree} = \SI{0.50}{\newton}$, roughly \SI{18}{\percent} of the robot's weight. During stance, this force acts at a moment arm of about \SI{1973}{\milli\meter}, which makes it sufficient for orientation control. Similar hybrid combinations of legs and aerodynamic actuators have been demonstrated on multimodal robots~\cite{kim2021bipedal, sihite2023multi} and hopping robots~\cite{haldane2017repetitive, zhu2022pogodrone, wang2024terrestrial, bai_agile_2024}.

We model the module as a planar system with lean angle $\psi$ from vertical and attitude error $\psi_e \triangleq \psi - \psi_\text{des}$ (\cref{fig:balancing_diagram}). The leg carries three lumped masses: the body at position $x$ from the foot, the leg with inertia $I_\text{leg}$ and center at $L/2$, and the module at the tip $L$. The total mass is $M = m_\text{body} + m_\text{leg} + m_\text{bal}$, and $F$ is the net thruster force.

\emph{Contact phase.} With the foot planted, the body at its lowest position $x_0$, and a tripod contacting the ground for support, the system is an inverted pendulum about the foot (\cref{fig:balancing_diagram}~(a)). The body \ac{com} lies at half the body height. The system \ac{com} $x_\text{COM}$ and the moment of inertia about the pivot $I_\text{piv}$ are
\begin{align}
    x_\text{COM} &= \frac{m_\text{body}\,x_0 + m_\text{leg}\,L/2 + m_\text{bal}\,L}{M},
    \label{eq:xcom_contact}\\
    I_\text{piv} &= m_\text{body}\,x_0^2 + m_\text{leg}\!\left(\frac{L}{2}\right)^{\!2} + m_\text{bal}\,L^2 + I_\text{leg}.
    \label{eq:Ipiv}
\end{align}
Taking moments about the foot gives $I_\text{piv}\,\ddot\psi = M g\,x_\text{COM}\sin\psi + F L + \tau_\text{tri}$, where $\tau_\text{tri}$ is the reaction torque from the tripod contact. With $k_g \triangleq M g\,x_\text{COM}$, the computed-torque law
\begin{equation}
    F = -\frac{1}{L}\left(k_g\sin\psi + K_p\,\psi_e + K_d\,\dot\psi_e\right)
    \label{eq:contact_law}
\end{equation}
cancels gravity, so the closed loop reduces to the PD error dynamics
\begin{equation}
    I_\text{piv}\,\ddot\psi_e + K_d\,\dot\psi_e + K_p\,\psi_e = \tau_\text{tri},
    \label{eq:contact_err}
\end{equation}
with $\tau_\text{tri}$ acting as a bounded disturbance.

\emph{Aerial phase.} After liftoff, the system rotates about its own center of mass, where gravity exerts no moment (\cref{fig:balancing_diagram}~(b)). Replacing $x_0$ with the body position $x$ in \eqref{eq:xcom_contact} gives $x_\text{COM}(x)$, and the inertia $I_\text{COM}$ about this point follows by the parallel-axis theorem. The thrust acts with a moment arm $L - x_\text{COM}$, giving the gravity-free rotation $I_\text{COM}\,\ddot\psi = F(L - x_\text{COM})$, which the PD law
\begin{equation}
    F = -\frac{1}{L - x_\text{COM}}\left(K_p\,\psi_e + K_d\,\dot\psi_e\right)
    \label{eq:air_law}
\end{equation}
reduces to the homogeneous form of \eqref{eq:contact_err} ($\tau_\text{tri} = 0$) with $I_\text{COM}$ for $I_\text{piv}$. Because $I_\text{COM} \ne I_\text{piv}$, the module switches from ground to aerial gains at liftoff, triggered by a signal from the body. The corrective torque scales with $L - x_\text{COM}$, which is short while the body is near the module and would grow if it retracted toward the foot.

The module holds the leg upright or at a commanded lean of up to \SI{15}{\degree}, and cuts all thrust past a \SI{60}{\degree} tilt as a safety measure. A separate PD loop regulates the module's rotation about the leg axis.

\subsection{Experiment Setup}\label{subsec:experiment_setup}
\begin{figure}[tpb]
        \centering
        \includegraphics[width=.6\columnwidth]{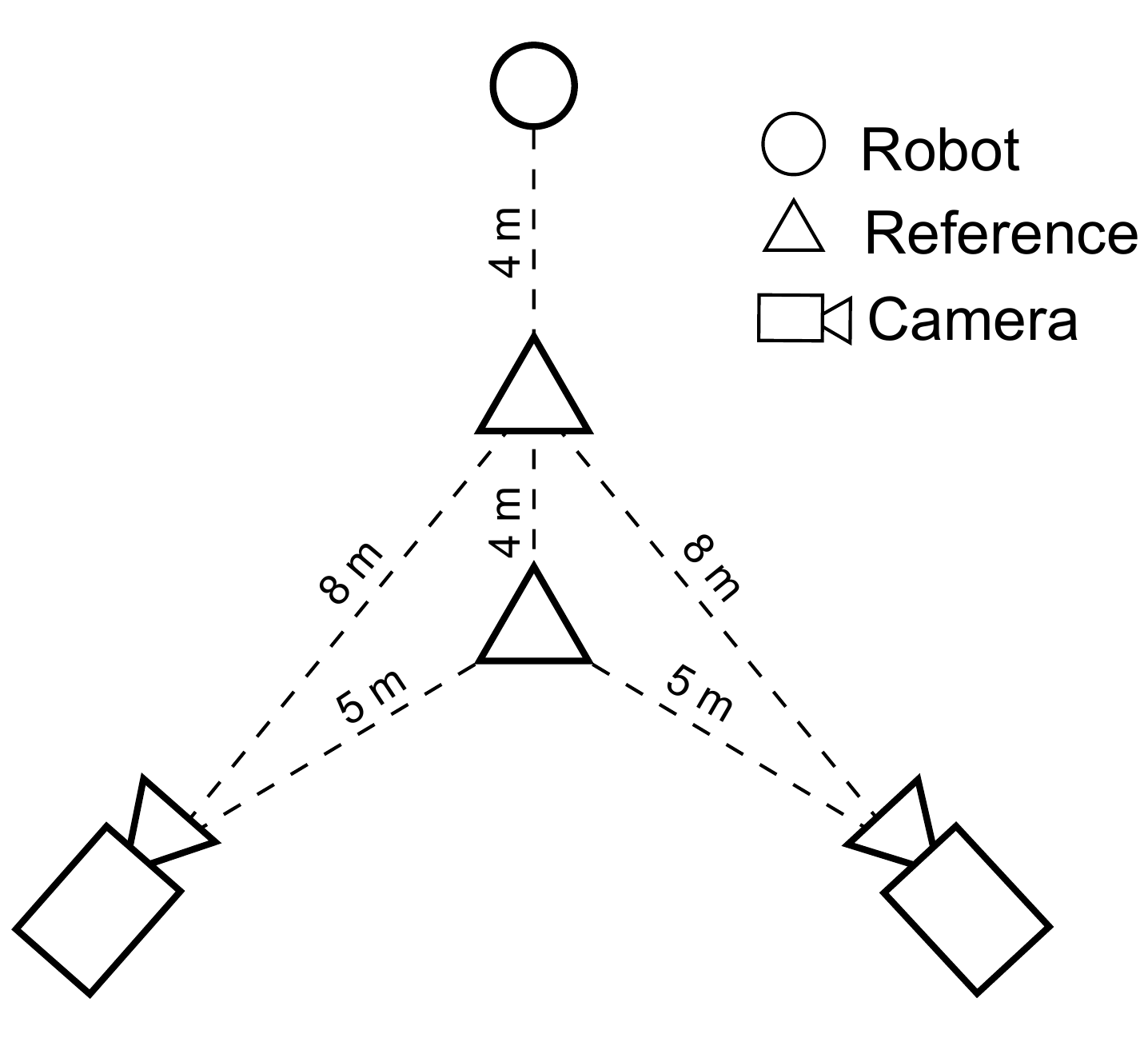}   
        \caption{Top-down layout of the vertical jump test setup, showing the placements of the two GoPro cameras, the meter stick length references, and the robot.}
        \vspace{-4mm}
        \label{fig:experiment_setup}
    \end{figure}

We conducted vertical jump tests in an open grassy field, using cameras and length references in a known geometry to measure the robot's motion in flight (\cref{fig:experiment_setup}). Two GoPro HERO13 Black cameras on tripods, shooting at 120~\ac{fps}, recorded the jump, with two meter sticks suspended on tripods as length references for calibrating the camera poses. The cameras, robot, and references were placed at the indicated positions, measured with a tape measure. The two GoPros were set to linear field of view mode to minimize lens distortion. Next to the right-hand one (\cref{fig:experiment_setup}), a cellphone camera at 30~\ac{fps} and a Sony Cybershot DSC-RX10 IV at 480~\ac{fps} also recorded the jump.

In each vertical jump, the robot was commanded to balance upright, then commanded to initiate its jump. Through the flight phase, the balancing module worked to maintain the upright attitude until touchdown.

In tilted jumps, the robot was commanded to balance upright, then the balance angle setpoint was moved to one side before the jump command. The balancing module was not commanded to maintain upright attitude after liftoff in tilted jumps.

\section{RESULTS}\label{subsec:results}
\subsection{Jump Height}
\begin{figure}[tpb]
        \centering
        \includegraphics[width=.5\columnwidth]{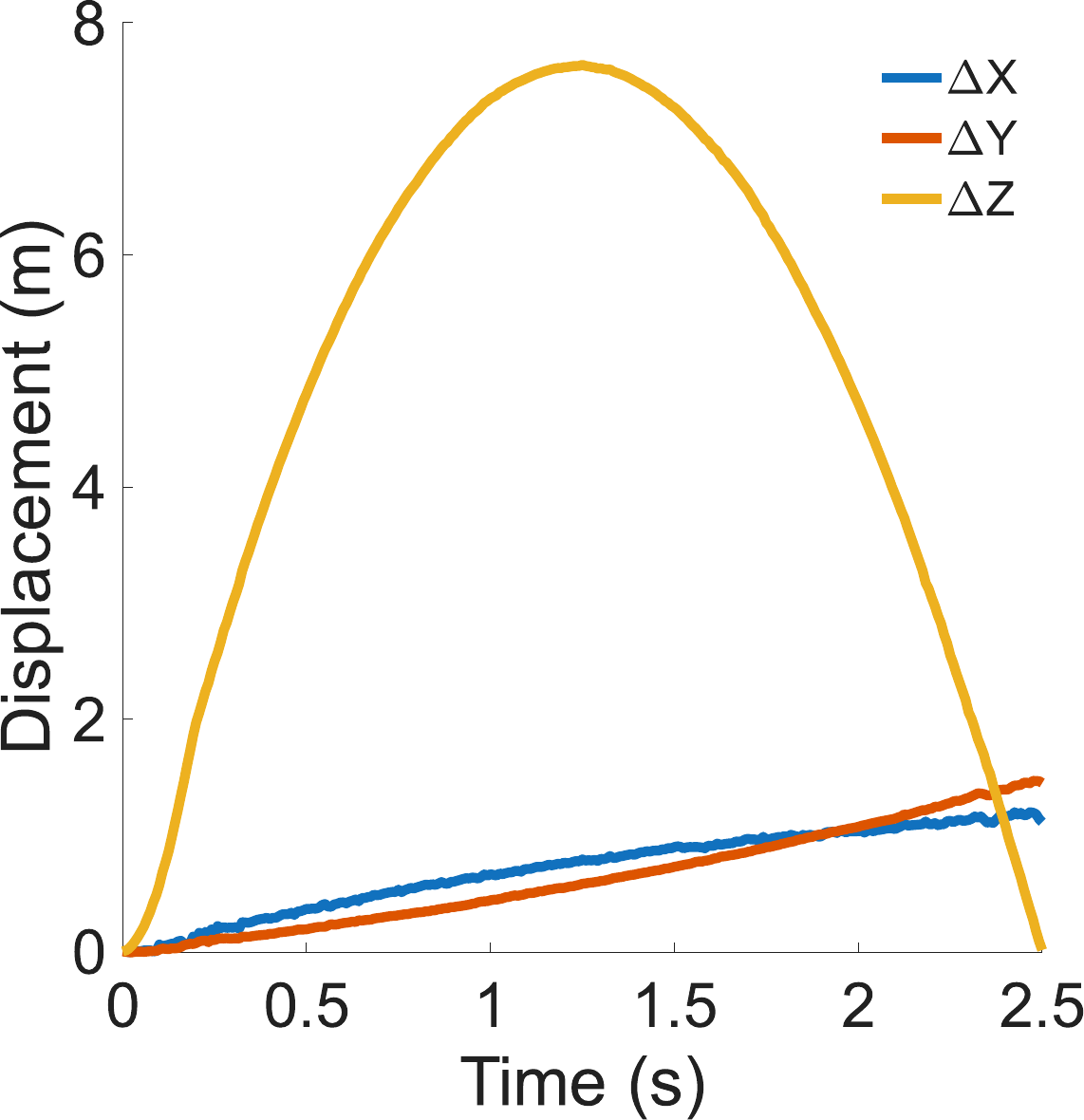}
        \caption{Body displacement during a single vertical jump, reconstructed by stereo triangulation. $\Delta Z$ is the vertical component; $\Delta X$ and $\Delta Y$ are horizontal. The body reached a peak height of \SI{7.6}{\meter}.}
        \vspace{-2mm}
    \label{fig:vertical_jump_trajectory}
    \end{figure}

We reconstructed the robot's jump trajectory in 3D by stereo triangulation from the two calibrated cameras of \cref{subsec:experiment_setup}, tracking the body from the start of the jump until touchdown. The resulting displacement is shown in \cref{fig:vertical_jump_trajectory}, with the $Z$ axis vertical. The body reached a maximum height $\Delta Z$ of approximately \SI{7.6}{\meter}, and the jump lasted about \SI{2.5}{\second}. The horizontal displacements $\Delta X$ and $\Delta Y$ each remained below \SI{1.6}{\meter}. To compute \ac{vja}, we take the \ac{com} height as that of the body. During flight, the body sits at the top of the leg, adjacent to the balancing module. The two dominant masses, the body (\SI{192}{\gram}) and the balancing module (\SI{55}{\gram}), are therefore nearly co-located. The leg tube (\SI{27}{\gram}) is the only mass distributed along the length, and its small share of the total \SI{281}{\gram} makes its offset negligible.

\begin{figure}[tpb]
        \centering
        \includegraphics[width=.95\columnwidth]{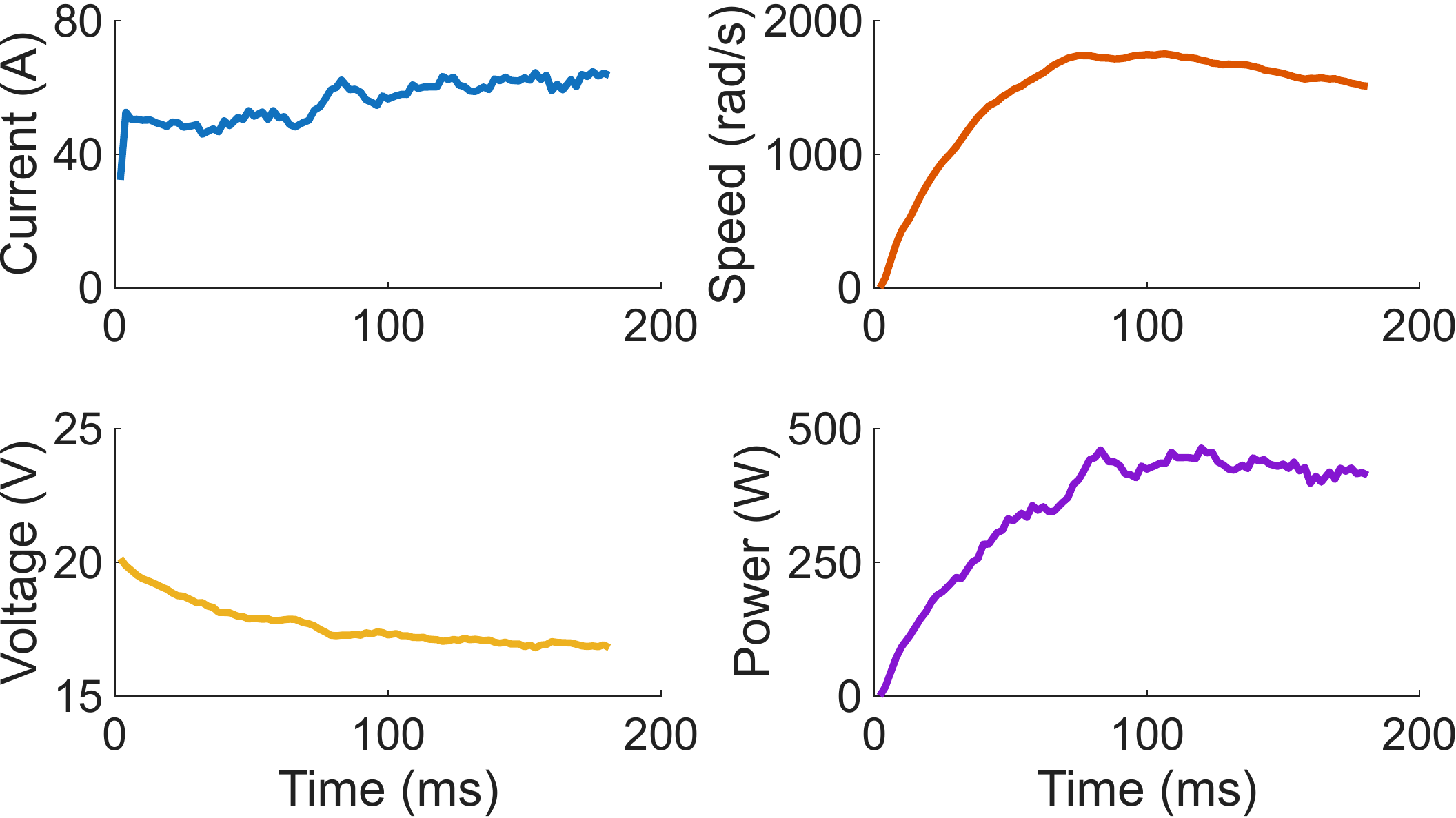}
        \caption{Main actuator telemetry during the jump: winding current, motor shaft velocity, bus voltage, and mechanical power. After an initial ramp, the power holds near its peak for the rest of the stroke.}
        \vspace{-4mm}
    \label{fig:body_motor_log}
    \end{figure}

The main actuator telemetry is shown in \cref{fig:body_motor_log}. We computed the mechanical output power from $P = K_t I \dot\theta$, where $I$ and $\dot\theta$ are the measured winding current and motor speed, and $K_t = \SI{0.0043}{\newton\meter\per\ampere}$ is the torque constant from the manufacturer's datasheet. Following an initial ramp, the power settles near its peak and remains there for the rest of the stroke. This result is the intended effect of the fabric-wrap transmission, as described in \cref{subsec:fabric_wrap_transmission}.

\subsection{Attitude Control}
\begin{figure}[tpb]
        \centering
        \includegraphics[width=.8\columnwidth]{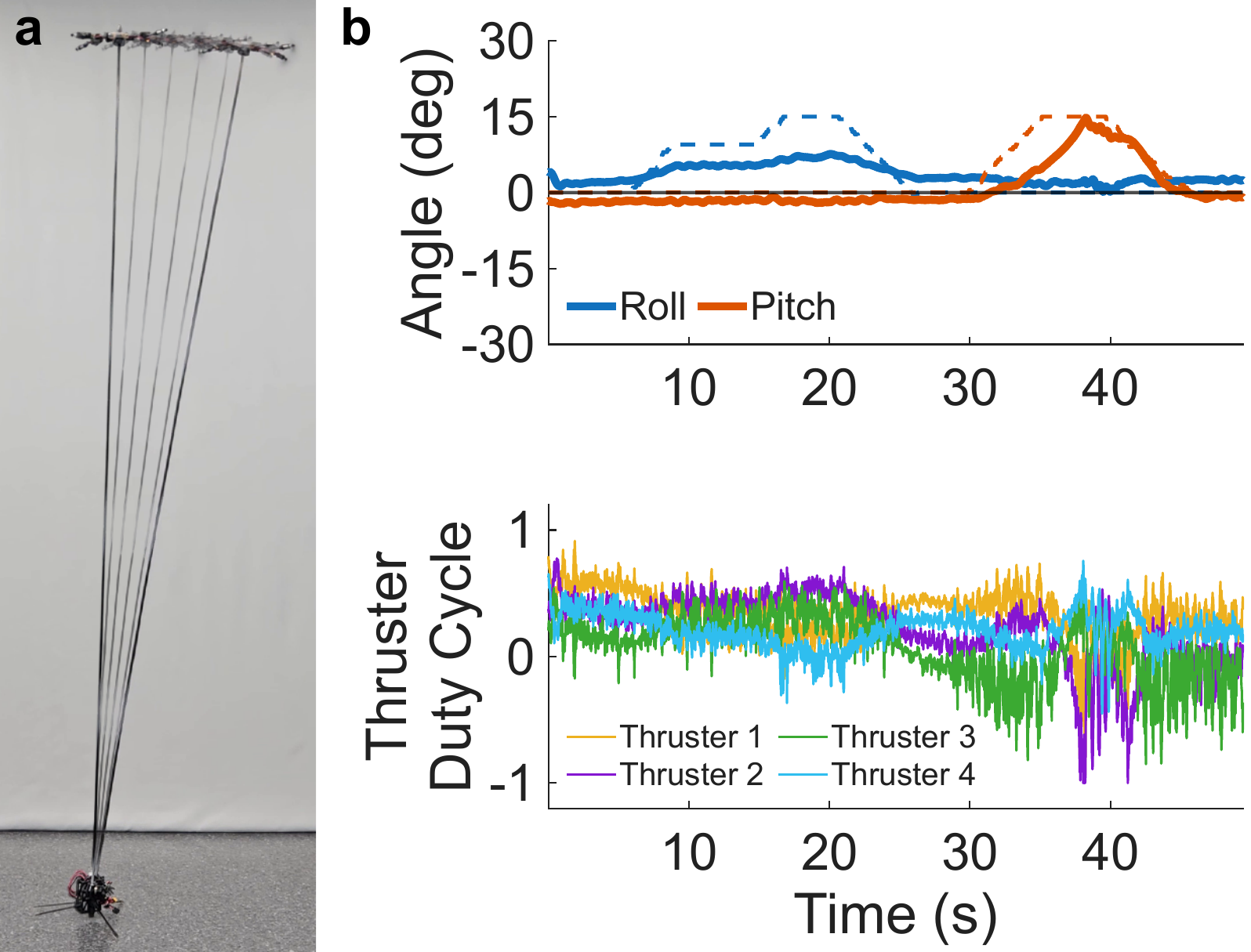}   
        \caption{On-ground attitude control. (a)~Overlay of the robot setting a pitch lean angle. (b)~Commanded (dashed) and measured (solid) roll and pitch, with the thruster duty cycle below.}
        \vspace{-2mm}
    \label{fig:tilting}
    \end{figure}

The balancing module controls the robot's attitude in both the stance and flight phases. During stance, it rejects disturbances such as wind to keep the leg upright, and it sets the lean angle from which the robot jumps. In flight, it works to stabilize the attitude. \Cref{fig:tilting} shows the attitude control on the ground. By holding a commanded lean, the robot can tilt before a jump to set its takeoff direction (\cref{fig:tilting}~(a)). \Cref{fig:tilting}~(b) shows that the measured angles follow the commands, as the roll and then the pitch setpoint are stepped to \SI{15}{\degree} and returned to level.

\begin{figure}[tpb]
        \centering
        \includegraphics[width=.7\columnwidth]{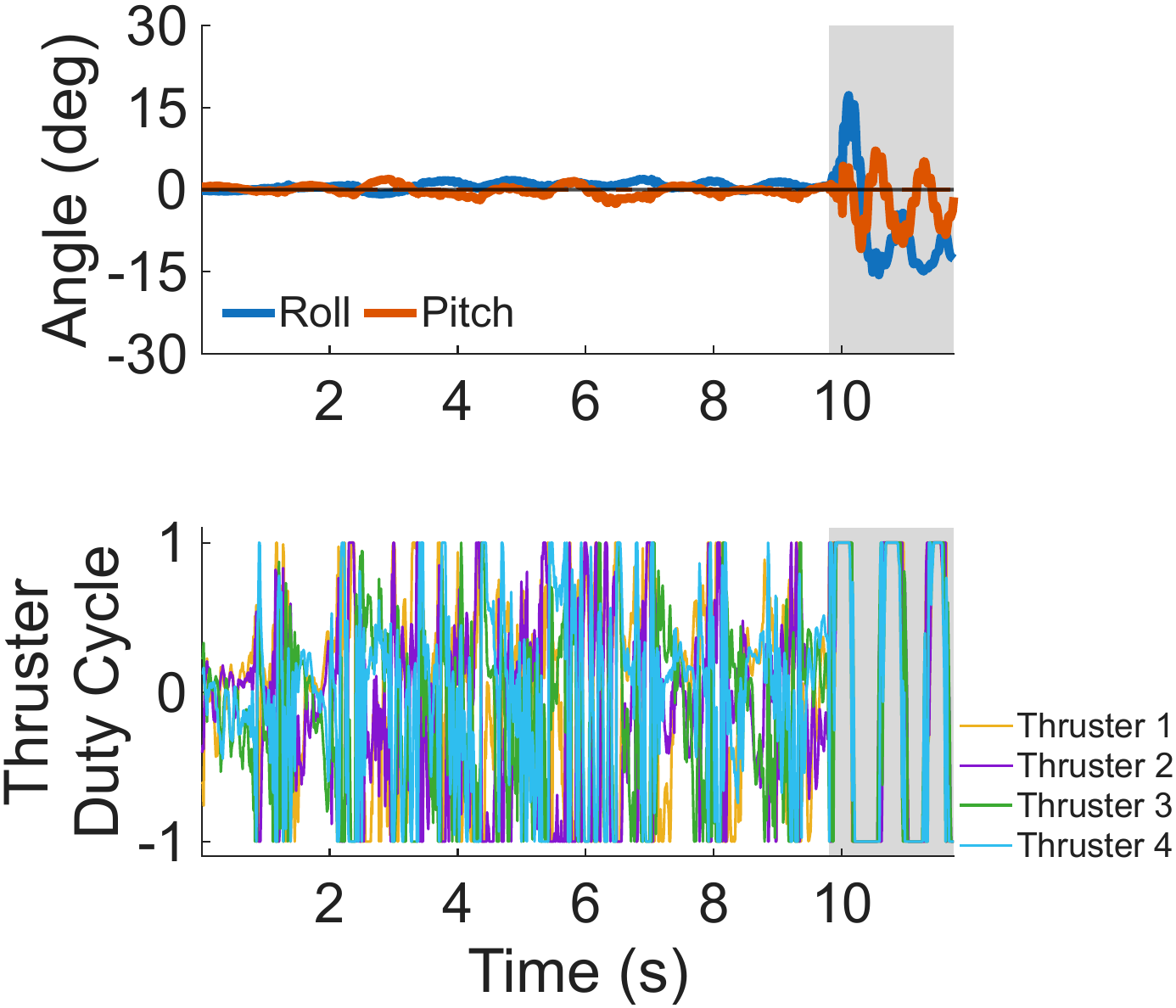}   
        \caption{Roll and pitch telemetry of the balancing module during a vertical jump (top); the gray region marks the post-liftoff phase. The controller switches from ground to aerial gains after liftoff. The bottom plot shows the thruster duty cycle.}
        \vspace{-4mm}
        \label{fig:vertical_jump_balancer_log}
    \end{figure}

\Cref{fig:vertical_jump_balancer_log} shows the telemetry of the balancing module during a vertical jump. The desired angle was zero throughout, and the controller held roll and pitch within a few degrees of level during stance. At liftoff it switched from ground to aerial gains. The balancing module reduced the aerial attitude deviations but did not fully settle them. As discussed in \cref{subsec:balancing_module}, exploiting the long leg's moment arm requires the body to move away from the balancing module. After liftoff, the \ac{com} stays close to the thrusters, since retracting the body toward the foot in flight is not yet implemented.

\subsection{Tilted Jump}
The robot can steer its jump direction during stance by leaning the leg with the balancing module. This potentially lets the robot place its touchdown point over successive jumps, while the main actuator sets the jump height independently.

\begin{figure}[tbp]
        \centering
        \includegraphics[width=.8\columnwidth]{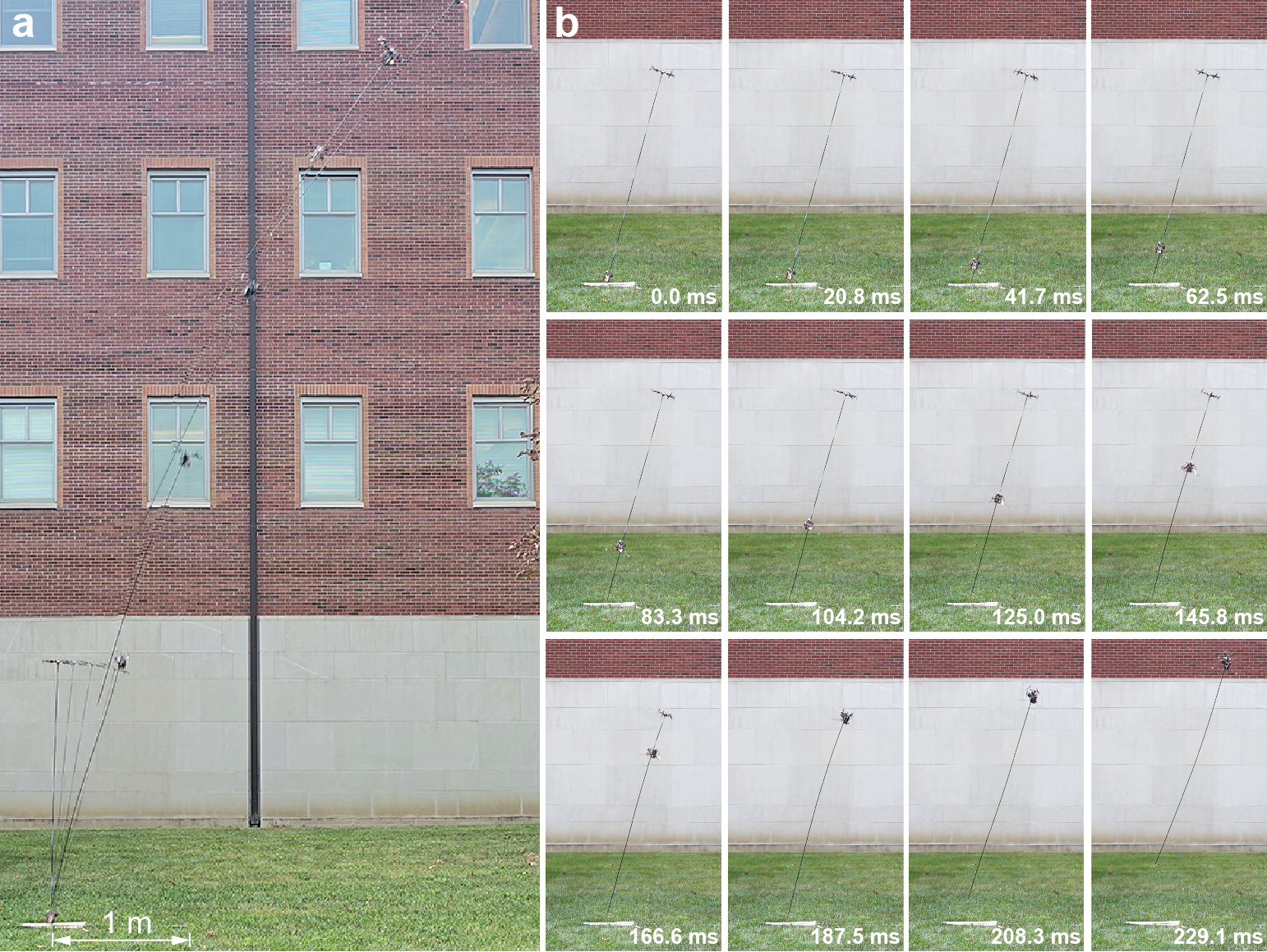}
        \caption{Tilted jump. (a)~Overlay of the lean and jump, with frames \SI{166.7}{\milli\second} apart after liftoff; the reference stick on the ground is \SI{1}{\meter} long. (b)~Snapshot sequence of the stance and liftoff phase at \SI{20.8}{\milli\second} intervals.}
        \vspace{-2mm}
        \label{fig:tilted_jump_overlay}
    \end{figure}

\begin{figure}[tbp]
        \centering
        \includegraphics[width=.6\columnwidth]{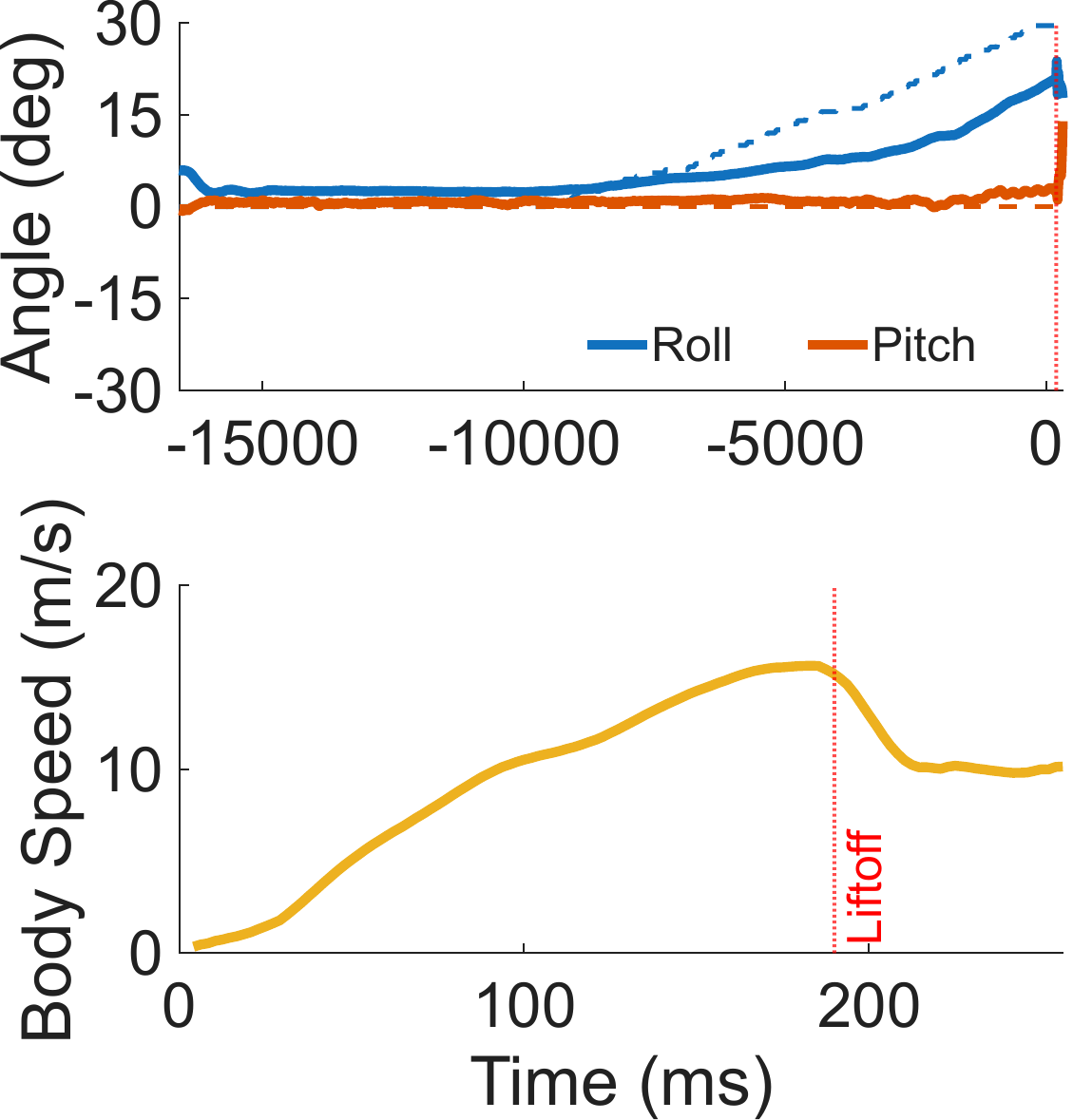}
        \caption{Balancing-module telemetry (top): solid lines are measured values, dashed lines are setpoints. The robot balances, tilts, and then jumps. The bottom plot shows the body speed following the jump command issued at \SI{0}{\second}; the robot lifts off at approximately \SI{190}{\milli\second}.}
        \vspace{-4mm}
        \label{fig:tilted_jump_plot}
    \end{figure}

We conducted tilted jump tests outdoors, shown in \cref{fig:tilted_jump_overlay}. The robot first balances upright and then leans as the setpoint changes. Once it reaches a lean of about \SI{20}{\degree}, the jump is commanded, and the robot lifts off approximately \SI{190}{\milli\second} later. \Cref{fig:tilted_jump_plot} shows the balancing-module telemetry over the whole sequence, together with the body's translational speed after the jump is initiated.

For this jump, the module was not commanded to stabilize attitude in the air. At liftoff, the robot is already leaning near the limit the module can recover from in flight, since the body cannot reposition along the leg to lengthen the moment arm (\cref{subsec:balancing_module}). Aerial stabilization was therefore not attempted. This tilted jump is a step toward directional multi-jump locomotion, where controlling the takeoff direction on each hop would let the robot traverse a sequence of targets.

\section{CONCLUSION AND FUTURE WORK}
A high-power \ac{dd} leg can enable high and rapid jumping motions.
The robot achieved \SI{7.6}{\meter} high jumps, the highest yet demonstrated by a \ac{dd} electrically actuated robot without mechanical energy storage elements like springs. Its high power-to-mass ratio also enabled it to achieve the highest vertical jumping agility at \SI{5.30}{\meter\per\second} among all electrically actuated robots of any transmission type. Expanding the envelope of jumping robot height and power enables larger, faster motions to clear large obstacles and make rapid maneuvers. Similar robot designs could even produce larger jumps on other planetary bodies like Earth's Moon, where surface gravity is lower, allowing jumping robots to cross obstacles like crater rims too challenging for conventional rover designs or capture drone-like overhead images in an atmosphere-less environment where conventional drones cannot fly.

Powerful \ac{dd} actuation that can propel a jump in less than half a second may be particularly suited to hopping motion in which a subsequent jump is initiated as soon as the robot lands to produce a running or bouncing gait. While this hopping motion was not demonstrated in this work, we validated the fabric-wrap transmission power output and jumping performance, balancing module control authority on the ground and in the air, and demonstrated that all components have sufficient durability to survive jumps and landings outdoors. Future work will develop reliable methods to reposition the body in the air for immediate jumps after landing, and introduce control techniques to enable continuous hopping like that of prior work~\cite{haldane2017repetitive, bai_agile_2024}. 

Propulsion with the fabric-wrap transmission may be suitable for other behaviors requiring fast launches and a low system mass like self-launching fixed-wing drones. Future work can explore other applications of this propulsion strategy and the effect of scaling the design to higher mass systems or different length scales, either longer or shorter.

\section*{Acknowledgment}
The authors would like to thank Vasil Iakimovitch and Grace Carsello for helping with the outdoor tests.





\bibliographystyle{IEEEtran}
\bibliography{LJL2026}

@article{haldane_robotic_2016,
	title = {Robotic vertical jumping agility via series-elastic power modulation},
	volume = {1},
	issn = {2470-9476},
	url = {https://www.science.org/doi/10.1126/scirobotics.aag2048},
	doi = {10.1126/scirobotics.aag2048},
	language = {en},
	number = {1},
	urldate = {2025-02-18},
	journal = {Science Robotics},
	author = {Haldane, Duncan W. and Plecnik, M. M. and Yim, J. K. and Fearing, R. S.},
	month = dec,
	year = {2016},
	pages = {eaag2048},
}

@article{hawkes_engineered_2022,
	title = {Engineered jumpers overcome biological limits via work multiplication},
	volume = {604},
	issn = {0028-0836, 1476-4687},
	url = {https://www.nature.com/articles/s41586-022-04606-3},
	doi = {10.1038/s41586-022-04606-3},
	language = {en},
	number = {7907},
	urldate = {2025-02-18},
	journal = {Nature},
	author = {Hawkes, Elliot W. and Xiao, Charles and Peloquin, Richard-Alexandre and Keeley, Christopher and Begley, Matthew R. and Pope, Morgan T. and Niemeyer, Günter},
	month = apr,
	year = {2022},
	pages = {657--661},
}

@inproceedings{kau_stanford_2019,
	address = {Montreal, QC, Canada},
	title = {Stanford {Doggo}: {An} {Open}-{Source}, {Quasi}-{Direct}-{Drive} {Quadruped}},
	copyright = {https://ieeexplore.ieee.org/Xplorehelp/downloads/license-information/IEEE.html},
	isbn = {978-1-5386-6027-0},
	shorttitle = {Stanford {Doggo}},
	url = {https://ieeexplore.ieee.org/document/8794436/},
	doi = {10.1109/ICRA.2019.8794436},
	language = {en},
	urldate = {2025-02-18},
	booktitle = {2019 {International} {Conference} on {Robotics} and {Automation} ({ICRA})},
	publisher = {IEEE},
	author = {Kau, Nathan and Schultz, Aaron and Ferrante, Natalie and Slade, Patrick},
	month = may,
	year = {2019},
	pages = {6309--6315},
}

@inproceedings{kovac_miniature_2008,
	address = {Pasadena, CA, USA},
	title = {A miniature 7g jumping robot},
	isbn = {978-1-4244-1646-2},
	url = {http://ieeexplore.ieee.org/document/4543236/},
	doi = {10.1109/ROBOT.2008.4543236},
	language = {en},
	urldate = {2025-05-06},
	booktitle = {2008 {IEEE} {International} {Conference} on {Robotics} and {Automation}},
	publisher = {IEEE},
	author = {Kovac, Mirko and Fuchs, Martin and Guignard, Andre and Zufferey, Jean-Christophe and Floreano, Dario},
	month = may,
	year = {2008},
	pages = {373--378},
}

@article{kenneally_design_2016,
	title = {Design {Principles} for a {Family} of {Direct}-{Drive} {Legged} {Robots}},
	volume = {1},
	copyright = {https://ieeexplore.ieee.org/Xplorehelp/downloads/license-information/OAPA.html},
	issn = {2377-3766, 2377-3774},
	url = {http://ieeexplore.ieee.org/document/7403902/},
	doi = {10.1109/LRA.2016.2528294},
	language = {en},
	number = {2},
	urldate = {2025-05-06},
	journal = {IEEE Robotics and Automation Letters},
	author = {Kenneally, Gavin and De, Avik and Koditschek, D. E.},
	month = jul,
	year = {2016},
	pages = {900--907},
}

@article{jung_jumproach_2019,
	title = {{JumpRoACH}: {A} {Trajectory}-{Adjustable} {Integrated} {Jumping}–{Crawling} {Robot}},
	volume = {24},
	copyright = {https://ieeexplore.ieee.org/Xplorehelp/downloads/license-information/IEEE.html},
	issn = {1083-4435, 1941-014X},
	shorttitle = {{JumpRoACH}},
	url = {https://ieeexplore.ieee.org/document/8675459/},
	doi = {10.1109/TMECH.2019.2907743},
	language = {en},
	number = {3},
	urldate = {2025-05-11},
	journal = {IEEE/ASME Transactions on Mechatronics},
	author = {Jung, Gwang-Pil and Casarez, Carlos S. and Lee, Jongeun and Baek, Sang-Min and Yim, So-Jung and Chae, Soo-Hwan and Fearing, Ronald S. and Cho, Kyu-Jin},
	month = jun,
	year = {2019},
	pages = {947--958},
}

@article{bai_agile_2024,
	title = {An agile monopedal hopping quadcopter with synergistic hybrid locomotion},
	volume = {9},
	issn = {2470-9476},
	url = {https://www.science.org/doi/10.1126/scirobotics.adi8912},
	doi = {10.1126/scirobotics.adi8912},
	language = {en},
	number = {89},
	urldate = {2025-06-22},
	journal = {Science Robotics},
	author = {Bai, Songnan and Pan, Qiqi and Ding, Runze and Jia, Huaiyuan and Yang, Zhengbao and Chirarattananon, Pakpong},
	month = apr,
	year = {2024},
	pages = {eadi8912},
}

@inproceedings{haldane2017repetitive,
  title={Repetitive extreme-acceleration (14-g) spatial jumping with Salto-1P},
  author={Haldane, Duncan W and Yim, Justin K and Fearing, Ronald S},
  booktitle={2017 IEEE/RSJ International Conference on Intelligent Robots and Systems (IROS)},
  pages={3345--3351},
  year={2017},
  organization={IEEE}
}

@inproceedings{klemm_ascento_2019,
	address = {Montreal, QC, Canada},
	title = {Ascento: {A} {Two}-{Wheeled} {Jumping} {Robot}},
	copyright = {https://ieeexplore.ieee.org/Xplorehelp/downloads/license-information/IEEE.html},
	isbn = {978-1-5386-6027-0},
	shorttitle = {Ascento},
	url = {https://ieeexplore.ieee.org/document/8793792/},
	doi = {10.1109/ICRA.2019.8793792},
	language = {en},
	urldate = {2026-04-01},
	booktitle = {2019 {International} {Conference} on {Robotics} and {Automation} ({ICRA})},
	publisher = {IEEE},
	author = {Klemm, Victor and Morra, Alessandro and Salzmann, Ciro and Tschopp, Florian and Bodie, Karen and Gulich, Lionel and Kung, Nicola and Mannhart, Dominik and Pfister, Corentin and Vierneisel, Marcus and Weber, Florian and Deuber, Robin and Siegwart, Roland},
	month = may,
	year = {2019},
	pages = {7515--7521},
}

@article{bai_parallel-elastic_2026,
	title = {Parallel-{Elastic} {Actuation} with {Reactive} {Latch} {Elevates} {Robotic} {Hopping} {Performance}: {Jump} {Height} and {Continuity}},
	copyright = {https://ieeexplore.ieee.org/Xplorehelp/downloads/license-information/IEEE.html},
	issn = {1552-3098, 1941-0468},
	shorttitle = {Parallel-{Elastic} {Actuation} with {Reactive} {Latch} {Elevates} {Robotic} {Hopping} {Performance}},
	url = {https://ieeexplore.ieee.org/document/11553415/},
	doi = {10.1109/TRO.2026.3701316},
	language = {en},
	urldate = {2026-06-27},
	journal = {IEEE Transactions on Robotics},
	author = {Bai, Songnan and Ding, Runze and Li, Song and Jia, Ruihan and Wang, Ruobing and Zhang, Zhiyuan and Wang, Fangzheng and Chirarattananon, Pakpong},
	year = {2026},
	pages = {1--20},
}

@inproceedings{suzuki_ramiel_2022,
	address = {Kyoto, Japan},
	title = {{RAMIEL}: {A} {Parallel}-{Wire} {Driven} {Monopedal} {Robot} for {High} and {Continuous} {Jumping}},
	copyright = {https://doi.org/10.15223/policy-029},
	isbn = {978-1-6654-7927-1},
	shorttitle = {{RAMIEL}},
	url = {https://ieeexplore.ieee.org/document/9981963/},
	doi = {10.1109/IROS47612.2022.9981963},
	language = {en},
	urldate = {2026-07-24},
	booktitle = {2022 {IEEE}/{RSJ} {International} {Conference} on {Intelligent} {Robots} and {Systems} ({IROS})},
	publisher = {IEEE},
	author = {Suzuki, Temma and Toshimitsu, Yasunori and Nagamatsu, Yuya and Kawaharazuka, Kento and Miki, Akihiro and Ribayashi, Yoshimoto and Bando, Masahiro and Kojima, Kunio and Kakiuchi, Yohei and Okada, Kei and Inaba, Masayuki},
	month = oct,
	year = {2022},
	pages = {5017--5024},
}

@misc{wagner_underactuated_2026,
	title = {Underactuated multimodal jumping robot for extraterrestrial exploration},
	url = {http://arxiv.org/abs/2603.06525},
	doi = {10.48550/arXiv.2603.06525},
	language = {en},
	urldate = {2026-09-05},
	publisher = {arXiv},
	author = {Wagner, Neil R. and Yim, Justin K.},
	month = mar,
	year = {2026},
	note = {arXiv:2603.06525 [cs.RO]},
}

@inproceedings{batts_design_2016,
	address = {Stockholm, Sweden},
	title = {Design of a hopping mechanism using a voice coil actuator: {Linear} elastic actuator in parallel ({LEAP})},
	isbn = {978-1-4673-8026-3},
	shorttitle = {Design of a hopping mechanism using a voice coil actuator},
	url = {http://ieeexplore.ieee.org/document/7487191/},
	doi = {10.1109/ICRA.2016.7487191},
	language = {en},
	urldate = {2026-09-05},
	booktitle = {2016 {IEEE} {International} {Conference} on {Robotics} and {Automation} ({ICRA})},
	publisher = {IEEE},
	author = {Batts, Zachary and Kim, Joohyung and Yamane, Katsu},
	month = may,
	year = {2016},
	pages = {655--660},
}

@incollection{kulic_untethered_2017,
	address = {Cham},
	title = {Untethered {One}-{Legged} {Hopping} in {3D} {Using} {Linear} {Elastic} {Actuator} in {Parallel} ({LEAP})},
	volume = {1},
	copyright = {http://www.springer.com/tdm},
	isbn = {978-3-319-50114-7 978-3-319-50115-4},
	url = {http://link.springer.com/10.1007/978-3-319-50115-4_10},
	doi = {10.1007/978-3-319-50115-4_10},
	language = {en},
	urldate = {2026-09-05},
	booktitle = {2016 {International} {Symposium} on {Experimental} {Robotics}},
	publisher = {Springer International Publishing},
	author = {Batts, Zachary and Kim, Joohyung and Yamane, Katsu},
	editor = {Kulić, Dana and Nakamura, Yoshihiko and Khatib, Oussama and Venture, Gentiane},
	year = {2017},
	note = {Series Title: Springer Proceedings in Advanced Robotics},
	pages = {103--112},
}

@inproceedings{park_variable-speed_2015,
	address = {Seattle, WA, USA},
	title = {Variable-speed quadrupedal bounding using impulse planning: {Untethered} high-speed {3D} {Running} of {MIT} {Cheetah} 2},
	isbn = {978-1-4799-6923-4},
	shorttitle = {Variable-speed quadrupedal bounding using impulse planning},
	url = {http://ieeexplore.ieee.org/document/7139918/},
	doi = {10.1109/ICRA.2015.7139918},
	language = {en},
	urldate = {2026-09-05},
	booktitle = {2015 {IEEE} {International} {Conference} on {Robotics} and {Automation} ({ICRA})},
	publisher = {IEEE},
	author = {Park, Hae-Won and {Sangin Park} and Kim, Sangbae},
	month = may,
	year = {2015},
	pages = {5163--5170},
}

@article{park_jumping_2021,
	title = {Jumping over obstacles with {MIT} {Cheetah} 2},
	volume = {136},
	issn = {09218890},
	url = {https://linkinghub.elsevier.com/retrieve/pii/S0921889020305431},
	doi = {10.1016/j.robot.2020.103703},
	language = {en},
	urldate = {2026-09-05},
	journal = {Robotics and Autonomous Systems},
	author = {Park, Hae-Won and Wensing, Patrick M. and Kim, Sangbae},
	month = feb,
	year = {2021},
	pages = {103703},
}

@article{park_high-speed_2017,
	title = {High-speed bounding with the {MIT} {Cheetah} 2: {Control} design and experiments},
	volume = {36},
	issn = {0278-3649, 1741-3176},
	shorttitle = {High-speed bounding with the {MIT} {Cheetah} 2},
	url = {https://journals.sagepub.com/doi/10.1177/0278364917694244},
	doi = {10.1177/0278364917694244},
	language = {en},
	number = {2},
	urldate = {2026-09-05},
	journal = {The International Journal of Robotics Research},
	author = {Park, Hae-Won and Wensing, Patrick M and Kim, Sangbae},
	month = feb,
	year = {2017},
	pages = {167--192},
}

@article{guenther2016energy,
  title={Energy-efficient monopod running with a large payload based on open-loop parallel elastic actuation},
  author={Guenther, Fabian and Iida, Fumiya},
  journal={IEEE Transactions on Robotics},
  volume={33},
  number={1},
  pages={102--113},
  year={2016},
  publisher={IEEE}
}

@inproceedings{hougen2000miniature,
  title={A miniature robotic system for reconnaissance and surveillance},
  author={Hougen, Dean F and Benjaafar, Saifallah and Bonney, Jordan C and Budenske, John R and Dvorak, Mark and Gini, Maria and French, Howard and Krantz, Donald G and Li, Perry Y and Malver, Fred and others},
  booktitle={Proceedings 2000 ICRA. Millennium Conference. IEEE International Conference on Robotics and Automation. Symposia Proceedings (Cat. No. 00CH37065)},
  volume={1},
  pages={501--507},
  year={2000},
  organization={IEEE}
}

@misc{ackerman_boston_2012,
	title = {Boston {Dynamics} {Sand} {Flea} {Robot} {Demonstrates} {Astonishing} {Jumping} {Skills} - {IEEE} {Spectrum}},
	url = {https://spectrum.ieee.org/boston-dynamics-sand-flea-demonstrates-astonishing-jumping-skills},
	language = {en},
	urldate = {2026-09-06},
	journal = {IEEE Spectrum},
	publisher = {IEEE},
	author = {Ackerman, Evan},
	month = mar,
	year = {2012},
}

@inproceedings{xu2025pinto,
  title={Pinto: A latched spring actuated robot for jumping and perching},
  author={Xu, Christopher Y and Yan, Jack and Yim, Justin K},
  booktitle={2025 IEEE International Conference on Robotics and Automation (ICRA)},
  pages={6244--6251},
  year={2025},
  organization={IEEE}
}

@article{divi2020latch,
  title={Latch-based control of energy output in spring actuated systems},
  author={Divi, Sathvik and Ma, Xiaotian and Ilton, Mark and St Pierre, Ryan and Eslami, Babak and Patek, SN and Bergbreiter, Sarah},
  journal={Journal of the Royal Society Interface},
  volume={17},
  number={168},
  pages={20200070},
  year={2020}
}

@article{hockman2017design,
  title={Design, control, and experimentation of internally-actuated rovers for the exploration of low-gravity planetary bodies},
  author={Hockman, Benjamin J and Frick, Andreas and Reid, Robert G and Nesnas, Issa AD and Pavone, Marco},
  journal={Journal of Field Robotics},
  volume={34},
  number={1},
  pages={5--24},
  year={2017},
  publisher={Wiley Online Library}
}

@inproceedings{hockman2022gravity,
  title={Gravity poppers: Hopping probes for the interior mapping of small solar system bodies},
  author={Hockman, Benjamin and Villa, Jacopo and French, Andrew and Chesley, Steven and Scheeres, Daniel J and McMahon, Jay},
  booktitle={2022 IEEE Aerospace Conference (AERO)},
  pages={1--26},
  year={2022},
  organization={IEEE}
}

@article{sihite2023multi,
  title={Multi-Modal Mobility Morphobot (M4) with appendage repurposing for locomotion plasticity enhancement},
  author={Sihite, Eric and Kalantari, Arash and Nemovi, Reza and Ramezani, Alireza and Gharib, Morteza},
  journal={Nature communications},
  volume={14},
  number={1},
  pages={3323},
  year={2023},
  publisher={Nature Publishing Group UK London}
}

@article{kim2021bipedal,
  title={A bipedal walking robot that can fly, slackline, and skateboard},
  author={Kim, Kyunam and Spieler, Patrick and Lupu, Elena-Sorina and Ramezani, Alireza and Chung, Soon-Jo},
  journal={Science Robotics},
  volume={6},
  number={59},
  pages={eabf8136},
  year={2021},
  publisher={American Association for the Advancement of Science}
}

@inproceedings{zhu2022pogodrone,
  title={Pogodrone: Design, model, and control of a jumping quadrotor},
  author={Zhu, Brian and Xu, Jiawei and Charway, Andrew and Salda{\~n}a, David},
  booktitle={2022 International Conference on Robotics and Automation (ICRA)},
  pages={2031--2037},
  year={2022},
  organization={IEEE}
}

@inproceedings{wang2024terrestrial,
  title={Terrestrial locomotion of pogox: From hardware design to energy shaping and step-to-step dynamics based control},
  author={Wang, Yi and Kang, Jiarong and Chen, Zhiheng and Xiong, Xiaobin},
  booktitle={2024 IEEE International Conference on Robotics and Automation (ICRA)},
  pages={3419--3425},
  year={2024},
  organization={IEEE}
}

@inproceedings{stoeter2002autonomous,
  title={Autonomous stair-hopping with scout robots},
  author={Stoeter, Sascha A and Rybski, Paul E and Gini, Maria and Papanikolopoulos, Nikos},
  booktitle={IEEE/RSJ international conference on intelligent robots and systems},
  volume={1},
  pages={721--726},
  year={2002},
  organization={IEEE}
}

@inproceedings{yang2025agile,
  title={Agile continuous jumping in discontinuous terrains},
  author={Yang, Yuxiang and Shi, Guanya and Lin, Changyi and Meng, Xiangyun and Scalise, Rosario and Castro, Mateo Guaman and Yu, Wenhao and Zhang, Tingnan and Zhao, Ding and Tan, Jie and others},
  booktitle={2025 IEEE International Conference on Robotics and Automation (ICRA)},
  pages={10245--10252},
  year={2025},
  organization={IEEE}
}
\end{document}